# Towards Explainable Khmer Polarity Classification

**Marry Kong**[†]     **Rina Buoy**[†]     **Sovisal Chenda**[†]     **Nguonly Taing**[†]

[†]Techo Startup Center, Ministry of Economy and Finance, Cambodia
Corresponding author: rina.buoy@techostartup.center

## Abstract

Khmer polarity classification is a fundamental natural language processing task that assigns a positive, negative, or neutral label to a given Khmer text input. Existing Khmer models typically predict the label without explaining the rationale behind the prediction. This paper proposes an explainable Khmer polarity classifier by fine-tuning an instruction-based reasoning Qwen-3 model. The notion of explainability in this paper is limited to self-explanations, which the model uses to rationalize its predictions. Experimental results show that the fine-tuned model not only predicts labels accurately but also provides reasoning by identifying polarity-related keywords or phrases to support its predictions. In addition, we contribute a new Khmer polarity dataset consisting of short- to medium-length casual, romanized, and mixed-code Khmer expressions. This dataset was constructed using both heuristic rules and human curation and is publicly available through a gated Hugging Face repository[1]. The fine-tuned Qwen-3 models are also made available in the same Hugging Face account.

**Keywords:** Polarity Classification, Explainability, LLM, Khmer Dataset.

## 1 Introduction

Polarity classification is a classical natural language processing task that assigns a discrete label to a given input text, typically from the set positive, negative, or neutral [1]. It is also commonly referred to as sentiment analysis. Fundamentally, polarity classification is a specific instance of general text classification or categorization, where the assigned label reflects sentiment [2]. Other examples of text classification tasks include spam detection, language identification, and authorship attribution [2].

Although Khmer is a low-resource language, the task of Khmer text classification has been relatively well studied. Various approaches, using both classical machine learning and deep learning, have been proposed for Khmer text classification. Similarly, several methods [1; 3; 4; 5; 6] for Khmer sentiment and polarity classification have been developed to assign sentiment labels to given Khmer text inputs. However, all of these methods are essentially black boxes, providing only the final label without any justification or explanation for the prediction. Understanding the model's reasoning is one of fundamental aspects of model explainability. It should be noted that the notion of explainability in this paper is limited to self-explanations, which the model uses to rationalize its predictions.

Thus, this paper proposes an explainable Khmer polarity classifier that not only predicts labels accurately but also provides justification or reasoning for its predictions. This is achieved by fine-tuning an instruction-based reasoning Qwen-3 [7] model. Experimental results demonstrate that the fine-tuned model delivers both accurate labels and coherent reasoning. Furthermore, to address the existing dataset gap, we introduce a new casual Khmer polarity dataset constructed through a combination of heuristic rules and human curation. The dataset comprises short- to medium-length casual, romanized, and mixed-code Khmer expressions. To the best of our knowledge, this is the first Khmer dataset of its kind. Our contributions are as follows:

1. We propose the first explainable Khmer polarity classifier that not only predicts labels accurately but also provides justification or reasoning for its predictions.

---

[1]rinabuoy/khmerpolarity_nonreasoning

2. We contribute a new Khmer polarity dataset consisting of short- to medium-length casual, romanized, and mixed-code Khmer expressions.

3. We make the resulting fine-tuned models publicly available to the community.

## 2 Related Work

### 2.1 Khmer Text Classification

Khmer is classified as a low-resource language due to limited research attention and the scarcity of datasets available for training and evaluation. Nonetheless, the task of Khmer text classification has been relatively well explored. Phann et al. [3] studied the classification of local news into nine categories using a range of classical machine learning methods, such as logistic regression, Naïve Bayes, and support vector machine (SVM) with term-frequency inverse document-frequency (TF–IDF) features. The SVM model with an radial basis function (RBF) kernel achieved the best performance. Jiang et al. [8] employed pretrained Khmer language models, including BERT [9] and ELECTRA [10], for Khmer news classification, reporting a top accuracy of 70.6% across eight news categories. Rifat and Imran [5] proposed a similar pretraining and fine-tuning approach using BERT for both sentiment and news classification, significantly outperforming the baseline FastText-based model. Similarly, Buoy et al. [4] utilized pretrained word embeddings (FastText) together with neural architectures such as recurrent neural networks (RNNs) and convolutional neural networks (CNNs) for Khmer text classification. Compared with a baseline pipeline using TF–IDF features and an SVM classifier, their models achieved improved accuracies on both multiclass and multilabel classification tasks.

In the area of sentiment detection, Prom et al. [6] proposed a method that combines BERT-based contextual features with a bidirectional long short-term memory (BiLSTM) classifier. Their approach achieved an accuracy of 86%, outperforming traditional machine learning algorithms in classifying Khmer text sentiments into negative, neutral, and positive categories. Similarly, Ye et al. [1] constructed a new Khmer polarity dataset and experimented with a range of classical machine learning techniques and text feature extractors.

### 2.2 Prompt-Based Sentiment Classification

Instruction-tuned large language models can be prompted to perform sentiment classification using zero-shot or few-shot prompting techniques, without the need for fine-tuning or gradient updates [11]. Nonetheless, for low-resource languages such as Khmer, their performance often suffers from hallucination and does not consistently align with user expectations. A prompt-based polarity detection was proposed for Czech language, using both zero-shot and few-shot prompting techniques [12]. The authors highlighted that the prompt-based method outperformed traditional finetuning especially on a limited training dataset. Also, the authors suggested that additional pretraining on target domain can enhance performance in a zero-shot prompting case.

### 2.3 Parameter-Efficient Fine-tuning

Fine-tuning a large model such as Qwen-3 by updating all parameters requires a substantial memory footprint. To address this, parameter-efficient fine-tuning approaches are recommended, especially in resource-constrained environments (e.g., a single GPU). Among these, the low-rank adaptation (LoRA) method [13] is one of the most widely used. LoRA approximates linear layer matrices with low-rank matrices containing fewer trainable parameters. Instead of updating all parameters, LoRA injects only trainable low-rank updates, reducing the number of trainable parameters by approximately 90% while maintaining performance. QLoRA [14] further reduces memory usage by using 4-bit quantized pretrained models.

## 3 Proposed Solution

This paper proposes an explainable Khmer polarity classifier. In this context, explainable means that the classifier not only predicts labels accurately but also provides justification or reasoning for its predictions. This notion is also known as self-explanation. This is achieved by fine-tuning an instruction-tuned large language model (LLM) with reasoning capability. For this purpose, we adopted the instruction-tuned Qwen3 models, which unify both a thinking mode for complex tasks and a non-thinking mode for simpler tasks within the same architecture. Due to resource constraints, we experimented only with the Qwen3-1.7B, Qwen3-4B, and Qwen3-8B variants, as these models are multilingual and support the Khmer language. Qwen-3 is the latest release of the Qwen model family, including Qwen-2 [15] and Qwen-2.5[16], with enhanced multilingual (Khmer included) instruction understanding and translation.

To fine-tune an instruction-tuned Qwen3 model into an explainable Khmer polarity classifier, we design a prompt that activates the model's thinking mode by guiding it to identify polarity-related keywords or phrases before predicting the final polarity label. These identified keywords or phrases serve as the basis for explaining the model's predictions. The reasoning prompt template is shown in Figure 1. As illustrated, the polarity-related keywords or phrases are enclosed between <think> and </think> tokens, which trigger the model to use this information as reasoning prior to concluding the final label during training.

When the polarity-related keywords or phrases are not available, the thinking mode is deactivated by applying the non-reasoning prompt template, as shown in Figure 2. This allows training to incorporate both datasets that include explicit polarity cues and those that do not, within the same model.

During inference, the thinking mode is always activated using the inference prompt template (Figure 3). Consequently, the model identifies any polarity-related keywords or phrases as part of its reasoning process before arriving at the final label. In this way, the model's predictions are inherently self-explainable.

We apply LoRA fine-tuning to reduce memory requirements. The LoRA configurations are shown in Table 1, and the number of LoRA trainable parameters, along with the total model parameters for each Qwen3 variant, are presented in Table 2. As provided in the table, the fine-tuning is limited to only the self-attention and feed-forward layers.

Table 1. LoRA configurations.

| Parameter | Value |
|-----------|-------|
| $r$ | 32 |
| $\alpha$ | 32 |
| dropout | 0 |
| bias | none |
| modules | projection layers |

Table 2. The LoRA trainable parameters. *Params*: model parameters.

| Model | LoRA Params | Full Params |
|-------|-------------|-------------|
| Qwen3-1B | 34M | 1.7B |
| Qwen3-4B | 66M | 4B |
| Qwen3-8B | 80M | 8B |

## 4 Datasets

### 4.1 Khmer Polarity (KP) Dataset

The dataset [1] comprises 10,000 manually labeled Khmer text inputs collected from online news articles. Each input is assigned one of three possible polarity labels: positive, negative, or neutral. The dataset is characterized by relatively long text inputs written in formal language. In addition, it includes keywords or phrases associated with the final label. However, the authors of this dataset used only the input texts and labels to train multiple Khmer polarity classifiers with various classical machine learning approaches, such as k-nearest neighbors (kNN) and support vector machines (SVM), and text feature extractors. As a result, their trained models lack explainability and justification. A few training samples from this dataset are shown in Table 3. For evaluation, the dataset is split into training (9,000) and test (1,000) sets.

### 4.2 Casual Khmer Polarity (CKP) Dataset

To complement the above KP dataset, we constructed a new casual Khmer polarity dataset [17] of approximately 16,500 texts,

```
    instruction = 'Classify the given text as positive, neutral, or negative:\n '
    conversations.append([
        {"role" : "user",      "content" :   instruction + {text}},
        {"role" : "assistant", "content" : f'<think> Because the input texcontains the
following {reasoning} </think>\n'+ label},
    ])
```

Figure 1. The reasoning prompt template for polarity classification. *text*: Khmer text input. *reasoning*: any polarity-related keywords or phrases. *label*: negative, positive, or neutral.

```
instruction = 'Classify the given text as positive, neutral, or negative:\n '
conversations.append([
    {"role" : "user",      "content" :   instruction + {text}},
    {"role" : "assistant", "content" : f'<think>\n\n</think>\n'+ label},
])
```

Figure 2. The non-reasoning prompt template for polarity classification. *text*: Khmer text input. *label*: negative, positive, or neutral.

```
instruction = 'Classify the given text as positive, neutral, or negative:\n '
conversations.append([
    {"role" : "user",      "content" :   instruction + {text}},
    {"role" : "assistant", "content" : '<think>'},
])
```

Figure 3. The reasoning prompt template for polarity classification during inference. *text*: Khmer text input.

Table 3. A few samples from the KP dataset, showing the Khmer text inputs, polarity-related keywords or phrases, and labels.

| Input Text | Reasoning | Label |
|---|---|---|
| យុវនារីណាដែលកាន់សៀវភៅពេលបច្ចុប្បន្ន គឺនឹងក្លាយទៅជាម្តាយដ៏ល្អនាពេលអនាគតៗ (Any young woman who holds a book today will become a good mother in the future.) | ម្តាយដ៏ល្អ (good mother) | positive |
| ឧទាហរណ៍ មនុស្សថ្លង់ ឬពិបាកស្តាប់ អាចស្តាប់មិនឮថាគរបាលនិយាយអ្វីខ្លះ ។ (For example, people who are deaf or hard of hearing may not be able to hear what the police are saying.) | ពិបាកស្តាប់/ស្តាប់មិនឮ (inaudible) | negative |
| រដ្ឋាភិបាលកម្ពុជាបានប្រកាសថា កាស៊ីណូទាំងអស់នៅក្នុងប្រទេសនឹងត្រូវបានអនុញ្ញាតឱ្យដំណើរ ការឡើងវិញ ប៉ុន្តែមានតែជាម្ស៊ីនល្បែង ក្លោស និងម៉ាស៊ីនលេងល្បែងតែប៉ុណ្ណោះៗ (The Cambodian government has announced that all casinos in the country will be allowed to reopen, but only with slot machines and slot games.) | ប្រកាស (announce) | neutral |

focusing on casual Khmer texts that are often used on social media, such as Facebook. These casual texts include Khmer, romanized, and mix-coded expressions. Data cleaning was applied by converting all English text to lowercase, removing punctuation and irrelevant symbols, eliminating emojis and special characters to reduce noise, and correcting misspellings to handle informal or noisy language effectively. A heuristic method first labeled comments automatically using sentiment-related keywords, followed by manual review and correction to ensure accuracy and handle misclassifications caused by context or sarcasm. We adopted the same labeling scheme (i.e., positive, negative, and neutral). Nonetheless, this dataset does not provide any identified polarity keywords or phrases. A few training samples from this dataset is provided in Table 4. For evaluation, this dataset is split into train (14,850) and test (1,650) sets.

Thus, by leveraging the proposed reasoning and non-reasoning prompting templates together with the KP dataset (with reasoning) and the CKP dataset (without reasoning), we fine-tune an explainable Khmer polarity classifier that can self-explain its predictions.

## 5 Experiments and Results

### 5.1 Experimental Setup

We used the Unsloth[2] fine-tuning framework, which enables memory-efficient training and inference on a single GPU. In all experiments, training was conducted for two full epochs on the combined data (i.e., KP and CKP) using a linear learning rate schedule with an initial rate of $2 \times 10^{-4}$ and a batch size of eight. The maximum context length was set to 2,048. Gradient accumulation was set to four, resulting in an effective batch size of 32 per gradient update. A weight decay of 0.01 was applied. We used instruction-tuned 4-bit quantized Qwen-3 model weights. For computing resources, training was performed on a single NVIDIA L4 GPU provided by Google Colab.

---

[2]https://unsloth.ai/

### 5.2 Results and Discussion

In this section, we present the experimental results followed by key analyses. We begin with a quantitative assessment using a standard classification accuracy metric, and then provide a qualitative assessment of reasoning and explainability.

#### 5.2.1 Quantitative Assessment of Classification

Table 5 presents an accuracy comparison between our fine-tuned Qwen-3 models and existing classical machine learning methods on the KP dataset. As shown, our fine-tuned models with the enabled thinking mode achieved an accuracy of 84%, compared to 60% by the best-performing tuned SGD model. The same table shows that the default Qwen-3 models (i.e., without fine-tuning) achieve an accuracy of 0% as they are unable to sufficiently understand Khmer language. As shown in the same table, When the thinking mode was not enabled (i.e., baseline), the model performance significantly degraded. In summary, the findings highlight the robustness of our proposed fine-tuned models.

However, on the CKP dataset in Table 6, our fine-tuned models performed comparably to existing deep learning–based models (i.e., RNN and CNN). Specifically, our fine-tuned Qwen-8B model achieved a classification accuracy of 87%, compared to 88% by the CNN model. The model's slight under-performance compared to the dedicated CNN model is likely due to the fact that fine-tuning was applied only to the self-attention and feed-forward layers, while the embedding layers, which lack sufficient Khmer language understanding, were not updated. Like on the KP dataset, the default Qwen-3 models achieved an accuracy of 0% and without the thinking mode (i.e., baseline), the model performance significantly degraded.

In summary, across both datasets, our proposed fine-tuned models either outperformed or achieved comparable performance to existing approaches. Moreover, our models provide reasoning and justification for their predictions (see next Section 5.2.2), a feature absent in previous methods.

Table 4. A few samples from the CKP dataset, showing the Khmer text inputs, and labels.

| Input Text | Reasoning | Label |
|---|---|---|
| អ្នបនៅវៃងខ្នៃងគាត់ឆ្ងាញ់ញុំមបានងទៅអ្នុងហើយៗ | | |
| ( The food at his place is delicious. I've been there once.) | - | positive |
| លំម៉ោងមើលលវ៉ៃុតហើយ ។ | | |
| (It's time to watch again.) | - | neutral |
| អ្នកកំពង់សោមមធ្វើអត់ស្ណាតទៅៗ | | |
| (The people of Kampong Som cannot prepare it well.) | - | negative |

Table 5. Assessment of classification accuracy on the KP dataset. **bold**: highest. *italic*: second highest. Baseline: w/o thinking.

| Model | Accuracy |
|---|---|
| kNN (Bigram) [1] | 0.48 |
| Decision Tree [1] | 0.54 |
| Random Forest [1] | 0.60 |
| SVM [1] | 0.59 |
| SGD [1] | 0.58 |
| SGD Tuning [1] | 0.60 |
| Qwen-1.7B (w/o FT) | 0.00 |
| Qwen-4B (w/o FT) | 0.00 |
| Qwen-8B (w/o FT) | 0.00 |
| Qwen-1.7B (Baseline) | 0.77 |
| Qwen-4B (Baseline) | 0.74 |
| Qwen-8B (Baseline) | *0.78* |
| Qwen-1.7B (Ours) | **0.84** |
| Qwen-4B (Ours) | **0.84** |
| Qwen-8B (Ours) | **0.84** |

Table 6. Assessment of classification accuracy on the CKP dataset. **bold**: highest. *italic*: second highest. Params : trainable only. Baseline: w/o thinking. FT: fine-tuning.

| Model | Params | Accuracy |
|---|---|---|
| XGBoost [17] | NA | 0.85 |
| RNN [17] | 3.91M | 0.87 |
| CNN [17] | 2.97M | **0.88** |
| Qwen-1.7B (w/o FT) | 0M | 0.00 |
| Qwen-4B (w/o FT) | 0M | 0.00 |
| Qwen-8B (w/o FT) | 0M | 0.00 |
| Qwen-1.7B (Baseline) | 34M | 0.81 |
| Qwen-4B (Baseline) | 66M | 0.85 |
| Qwen-8B (Baseline) | 80M | 0.82 |
| Qwen-1.7B (Ours) | 34M | 0.83 |
| Qwen-4B (Ours) | 66M | 0.86 |
| Qwen-8B (Ours) | 80M | *0.87* |

## 5.2.2 Qualitative Assessment of Reasoning and Explainability

In this section, we present a qualitative assessment of the model's reasoning and explainability. As described in Section 3, our fine-tuned models are trained to identify polarity-related keywords or phrases in the input text before producing the final label prediction. Table 7 provides some example Khmer text inputs and English translations, the model's derived polarity-related key words or phrases, and the final labels. The table shows that the model can not only give the correct labels but also extracts some polarity-related key words. These polarity-related key words are the basis or reasoning to understand or support why the model predicts what it predicts.

## 6 Future Work

Future work should include the following tasks:

1. The fine-tuning experiments were limited to the Qwen-3 model family only. Thus, future work should include other Khmer-supporting LLMs, such as GPT-0SS [18] or Sea-Lion [19].

2. The assessment of model explainability and reasoning is qualitative. Thus, future work should include other evaluation metrics, such as LLM as a judge.

3. The accuracies on the CP and CKP datasets are still limited because of the limited data quality and quantity. Thus, future work should focus on constructing a large-scale, high-quality Khmer polarity dataset.

Table 7. Qualitative assessment of the model reasoning and explainability based on the fine-tuned Qwen-8B model.

| Input Text | Reasoning | Label |
|---|---|---|
| អ្នបនៅហ្នឹងឆ្ងាញ់ គ្រប់មុខបងជាពិសេសសសម្នកគួរ។ | អ្នបនៅហ្នឹងឆ្ងាញ់/សម្នកគួរ | positive |
| (The food here is delicious, especially the stir-fried soup.) | (dellicious food/stir-fried soup) | - |
| ស្បែកដើងខ្ញុំទិញឲ្យរូនខ្ញុំយស្រួលពាក់ណាស់ចាំ ទៅទិញថ្មៀតៗ។ | ស្រួលពាក់ណាស់ | positive |
| (I bought the shoes for my children and they are so comfortable to wear. I can't wait to buy more.) | (comfortable to wear) | - |
| ប្រទេសកម្ពុជាស្ថិតនៅក្នុងតំបន់អាស៊ីអាគ្នេយ៍។ | ស្ថិត | neutral |
| (Cambodia is located in Southeast Asia.) | (is located) | - |
| សេះនេះមានខ្មៅៗ។ | មាន | neutral |
| (This horse is black.) | (is) | - |
| អ្នបមិនសូវមានរស់ជាតិឆ្ងាញ់។ | មិនសូវមានរស់ជាតិឆ្ងាញ់ | negative |
| (The food is not very tasty.) | (not very tasty) | - |
| គាត់ចូលចិត្តនិយាយអាក្រក់ពីអ្នកផ្សេងៗ។ | ចូលចិត្តនិយាយអាក្រក់ | negative |
| (He likes to talk badly about others.) | (talk badly) | - |

## 7 Conclusion

We propose the first explainable Khmer polarity classifier, which not only predicts the final label but also provides reasoning to justify its predictions. In this way, it becomes possible to explain why the model predicts what it does. Experimental results show that the fine-tuned Qwen-3 models either outperform or perform comparably to existing approaches on the KP and CKP datasets, respectively, for the Khmer polarity classification task.

**Acknowledgment**

We would like to acknowledge Ms. Udam Sonita for her contribution to the construction of the CKP dataset.